\documentclass[12pt]{spieman}  
\usepackage{tocloft}
\usepackage{epsfig}
\usepackage{setspace,graphicx,subcaption,caption,capt-of,textcomp}
\usepackage{doi}
\usepackage{mathtools,amsmath,amsfonts,amssymb,	array, xfrac}
\newcolumntype{M}[1]{>{\centering\arraybackslash}m{#1}}
\newcolumntype{P}{>{\raggedright\arraybackslash}p{3cm}}
\newcolumntype{N}{>{\centering\arraybackslash}m{2.5cm}}
\usepackage{pdflscape}
\usepackage{afterpage}
\usepackage{booktabs, multirow, tablefootnote, ragged2e}
\usepackage{xtab, multicol, multirow, makecell}
\usepackage[linesnumbered,ruled,vlined,boxed,commentsnumbered]{algorithm2e}
\usepackage[table]{xcolor}
\newcolumntype{s}[1]{>{\columncolor[HTML]{AAACED}} l}
\usepackage{subfiles}
\usepackage{blindtext}
\usepackage{import}
\title{PixelBNN: Augmenting the PixelCNN with batch normalization and the presentation of a fast architecture for retinal vessel segmentation}
\author[a,*]{Henry A. Leopold}
\author[b]{Jeff Orchard}
\author[a]{John S. Zelek}
\author[a]{Vasudevan Lakshminarayanan}
\affil[a]{Department of Systems Design Engineering}
\affil[b]{David R. Cheriton School of Computer Science}
\affil[ ]{University of Waterloo, Waterloo, ON, Canada, N2L3G1}

\cftpagenumbersoff{figure}
\cftpagenumbersoff{table} 

\begin{document} 
\maketitle
\graphicspath{{media/}}
\begin{abstract} \label{sec:abstract}
Analysis of retinal fundus images is essential for eye-care physicians in the diagnosis, care and treatment of patients. Accurate fundus and/or retinal vessel maps give rise to longitudinal studies able to utilize multimedia image registration and disease/condition status measurements, as well as applications in surgery preparation and biometrics. The segmentation of retinal morphology has numerous applications in assessing ophthalmologic and cardiovascular disease pathologies. The early detection of many such conditions is often the most effective method for reducing patient risk. Computer aided segmentation of the vasculature has proven to be a challenge, mainly due to inconsistencies such as noise and variations in hue and brightness that can greatly reduce the quality of fundus images. This paper presents PixelBNN, a highly efficient deep method for automating the segmentation of fundus morphologies. The model was trained, tested and cross tested on the DRIVE, STARE and CHASE\_DB1 retinal vessel segmentation datasets. Performance was evaluated using G-mean, Mathews Correlation Coefficient and F1-score. The network was 8.5$\times$ faster than the current state-of-the-art at test time and performed comparatively well, considering a 5$\times$ to 19$\times$ reduction in information from resizing images during preprocessing.
\end{abstract}


{\noindent \footnotesize\textbf{*}Contact: Henry A. Leopold,  \linkable{hleopold@uwaterloo.ca} }

\section{Introduction} \label{sec:intro}
The segmentation of retinal morphology has numerous applications in assessing ophthalmologic and cardiovascular disease pathologies, such as Glaucoma and Diabetes \cite{hoover_2000}. Diabetic retinopathy (DR) is one of the main causes of blindness globally, the severity of which can be rapidly assessed based on retinal vascular structure \cite{autoseg}. Glaucoma, another major cause for global blindness, can be diagnosed based on the properties of the optic nerve head (ONH). Analysis of the ONH typically requires the removal of vasculature for computational methods. Similar analyses of other structures within the eye benefit from the removal of retinal vessels making the segmentation and subtraction of vasculature critical to many forms of fundus analysis. Direct assessment of vessel characteristics such as length, width, tortuosity and branching patterns can uncover abnormal growth patterns or other disease markers - such as the presence of aneurysms, which are used to evaluate the severity of numerous health conditions including diabetes, arteriosclerosis, hypertension, cardiovascular disease and stroke \cite{ricci_2007}. For these types of diseases, early detection is critical in minimizing the risk complications and vision loss in the case of DR, glaucoma and other conditions of the eye \cite{cree_2008}; early detection is often the most effective method for reducing patient risk through modifications to lifestyle, medication and acute monitoring \cite{fraz_2015}. Similarly, the same information - this time gleaned from youth, can be used as indicators in the prediction of those individuals' health later in life \cite{abramoff_2010}. 

Retinal vessel segmentation from fundus images plays a key role in computer aided retinal analyses, either in the assessment of the vessels themselves or in vessel removal prior the evaluation of other morphologies, such as the ONH and macula. For this reason, it has been the most crucial step of practically all non-deep computer based analyses of the fundus \cite{fraz_2012b}. Automated computer image analysis provides a robust alternative to direct ophthalmoscopy by a medical specialist, providing opportunities for more comprehensive analysis through techniques such as batch image analysis \cite{azzopardi_2015}. As such, much research has gone into automatically measuring retinal morphology, traditionally utilizing images captured via fundus cameras. However, automatic segmentation of the vasculature has proven to be a challenge, mainly due to inconsistencies such as noise or variations in hue and brightness, which can greatly reduce the quality of fundus images \cite{fraz_2012a}. Traditional retinal pathology and morphology segmentation techniques often evaluate the green channel of RGB fundus images, as it is believed to be the ``best'' channel for assessing vascular tissue and lesions, while the red and blue channels suffer low contrast and high noise \cite{soares_2006}. Unfortunately, variations in image quality and patient ethnicity often invalidate this belief in real world settings.

Accurate feature extraction from retinal fundus images is essential for eye-care specialists in the care and treatment of their patients. Unfortunately, experts are often inconsistent in diagnosing retinal health conditions resulting in unnecessary complications\cite{almazroa_2015}. The use of computer aided detection (CAD) methods are being utilized to quantify the disease state of the retina, however most traditional methods are unable to match the performance of clinicians. These systems under-perform due to variations in image properties and quality, resulting from the use of varying capture devices and the experience of the user \cite{fraz_2012a}. To properly build and train an algorithm for commercial settings would require extensive effort by clinicians in the labelling of each and every dataset - a feat that mitigates the value of CAD systems. Overcoming these challenges would giver rise to longitudinal studies able to utilize multi-modal image registration and disease/condition status measurements, as well make applications in surgery preparation and biometrics more viable \cite{fraz_2012a}. 

The emergence of deep learning methods has enabled the development of CAD systems with an unprecedented ability to generalize across datasets, overcoming the shortcoming of traditional or ``shallow" algorithms. Computational methods for image analysis are divided into supervised and unsupervised techniques. Prior deep learning, supervised methods encompassed pattern recognition algorithms, such as k-nearest neighbours, decision trees and support vector machines (SVMs). Examples of such methods in the segmentation of retinal vessels include \emph{2D Gabor wavelet and Bayesian classifiers} \cite{soares_2006},  \emph{line operators and SVMs} \cite{ricci_2007} and \emph{AdaBoost-based classifiers} \cite{lupascu_2010}. Supervised methods require training materials be prepared by an expert, traditionally limiting the application of shallow methods. Unsupervised techniques stimulate a response within the pixels of an image to determine class membership and do not require manual delineations. The majority of deep learning approaches fall into the supervised learning category, due to their dependence on ground truths during training. Often, \textit{unsupervised} deep learning techniques refer to unsupervised pretraining for improving network parameter initialization as well as some generative and adversarial methods. 

Deep learning overcomes shallow methods' inability to generalize across datasets through the random generation and selection of a series of increasingly dimensional feature abstractions from combinations of multiple non-linear transformations on a dataset \cite{bengio_2013}. Applications of these techniques for object recognition in images first appeared in 2006 during the MNIST digit image classification problem, of which convolutional neural networks (CNNs) currently hold the highest accuracy \cite{hinton_2006}. Like other deep neural networks (DNNs), CNNs are designed modularly with a series of layers selected to address different classification problems.  A layer is comprised of an input, output, size (number of ``neurons") and a varying number of parameters/hyper-parameters that govern its operation. The most common layers include convolutional layers, pooling/subsampling layers and fully connected layers.

In the case of retinal image analysis, deep algorithms utilize a binary system, learning to differentiate morphologies based on performance masks manually delineated from the images. The current limitation with most unsupervised methods is that they utilize a set of predefined linear kernels to convolve the images or templates that are sensitive to variations in image quality and fundus morphologies \cite{azzopardi_2015}. Deep learning approaches overcome these limitations, and have been shown to outperform shallow methods for screening and other tasks in diagnostic retinopathy \cite{lecun_2015,abramoff_2016}. A recent review chapter discusses many of these issues and related methodologies \cite{dlra}. 

This paper presents PixelBNN, a novel variation of PixelCNN\cite{oord_2016b} - a dense fully convolutional network (FCN), that takes a fundus image as the input and returns a binary segmentation mask of the same dimension. The network was trained on resized images, deviating from other state-of-th-art methods which use cropping. The network was able to evaluate test images in 0.0466s, 8.5$\times$ faster than the state-of-the-art. Section 2 discusses the method and network architecture. Section 3 describes the experimental design. The resulting network performance is described in Section 4. Lastly, Section 5 discusses the results, future work and then concludes the paper. 
\section{Methodology} \label{sec:methodology}
Deep learning methods for retinal segmentation are typically based on techniques which have been successfully applied to image segmentation in other fields, and often utilize stochastic gradient descent (SGD) to optimize the network \cite{lecun_2015}. Recent work into stochastic gradient-based optimization has incorporated adaptive estimates of lower-order moments, resulting in the \emph{Adam} optimization method, which is further described below \cite{kingma_2014}. Adam was first successfully applied to the problem of retinal vessel segmentation by the authors, laying the foundation for this work \cite{leopold_2016}. 

Herein, a fully-residual autoencoder batch normalization network (``PixelBNN'') is trained via a random sampling strategy whereby samples are randomly distorted from a training set of fundus images and fed into the model. PixelBNN utilizes gated residual convolutional and deconvolutional layers activated by concatenated rectifying linear units (CReLU), similar to PixelCNN\cite{oord_2016a, oord_2016b} and PixelCNN++\cite{salimans_2017}. PixelBNN differs from its predecessors in three areas: (1) varied convolutional filter streams, (2) gating strategy, and (3) introduction of batch normalization layers\cite{batch_norm} from which it draws its name. 

\subsection{Datasets}  \label{sec:dataset}
\subsubsection{DRIVE}  \label{sec:drive}
\begin{figure}[!bp]
	\centering
	\begin{subfigure}{0.32\textwidth}
		\includegraphics[width=0.97\linewidth]{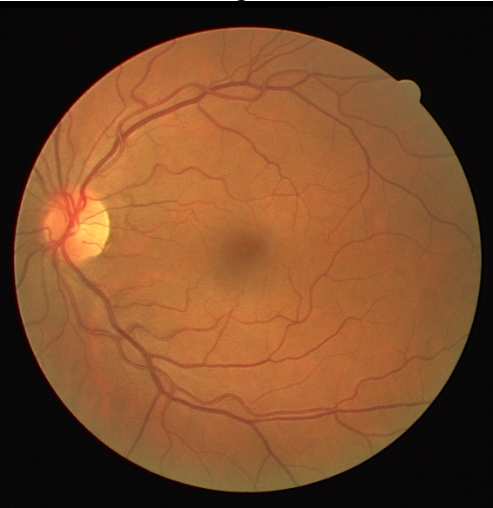} 
		\caption{}
		\label{fig:fundus_drive}
	\end{subfigure}
	\begin{subfigure}{0.32\textwidth}
		\includegraphics[width=0.97\linewidth]{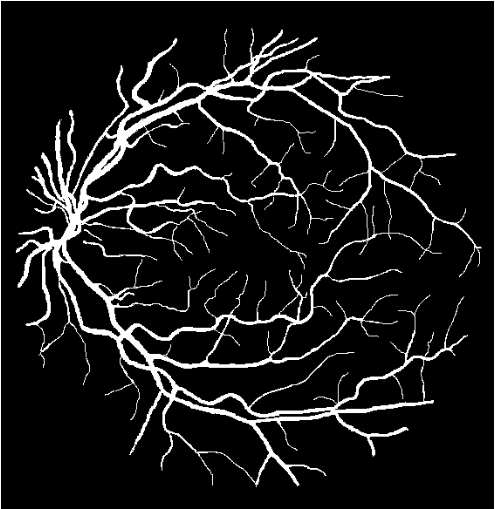}
		\caption{}
		\label{fig:vessel_gt}
	\end{subfigure}
	\begin{subfigure}{0.32\textwidth}
		\includegraphics[width=0.97\linewidth]{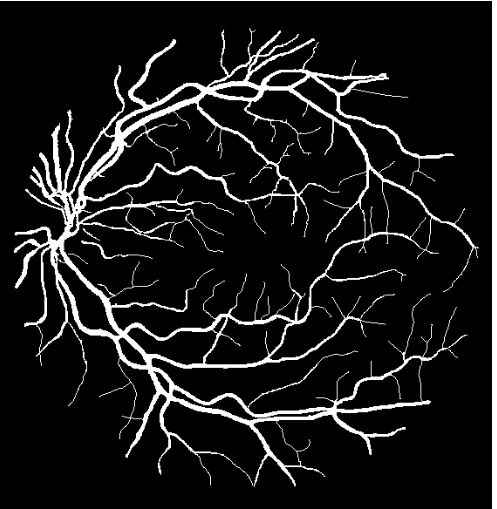} 
		\caption{}
		\label{fig:vessel_so}
	\end{subfigure}
	\caption{Sample set of the DRIVE dataset. (a): Fundus image. (b): First manual delineation, used as the ground truth. (c): Second manual delineation, referred to as the second human observer and used as the human performance benchmark. \cite{staal_2004}}
	\label{fig:drive}
\end{figure}

The CNN was trained and tested against the Digital Retinal Images for Vessel Extraction (DRIVE) database\footnote{http://www.isi.uu.nl/Research/Databases/DRIVE/}, a standardized set of fundus images used to gauge the effectiveness of classification algorithms \cite{staal_2004}. The images are 8 bits per RGBA channel with a 565$\times$584 pixel resolution. The data set comprises of 20 training images with manually delineated label masks and 20 test images with two sets of manually delineated label masks by the first and second human observers, as shown in Fig. \ref{fig:drive}. The images were collected for a diabetic retinopathy screening program in the Netherlands using a Canon CR5 non-mydriatic 3CCD camera with a 45{\textdegree} field of view \cite{staal_2004}. 

\subsubsection{STARE}  \label{sec:stare}
The Structured Analysis of the Retina database
\footnote{http://cecas.clemson.edu/{\raise.11ex\hbox{$\scriptstyle\mathtt{\sim}$}}ahoover/stare/}
 has 400 retinal images which are acquired using TopCon TRV-50 retinal camera with 35{\textdegree} field of view and pixel resolution of 700$\times$605. The database was populated and funded through the US National Institutes of Health \cite{hoover_2000}. A subset of the data is labelled by two experts, thereby providing 20 images with labels and ground truths. To compensate for the small number of images, four-fold cross validation was used. Therein, the network was trained over four runs, leaving five image out each time, resulting in all 20 images being evaluated without overlapping the training set, thusly minimizing network bias.

\subsubsection{CHASE\_DB1}  \label{sec:chase}
The third dataset used in this study is a subset of the Child Heart and Health Study in England database (CHASE\_DB1), containing 28 paired high-resolution (1280$\times$960 pixels) fundus images from each eye of 14 children, captured with a 30{\textdegree} field of view using a Nidek NM-200-D fundus camera. Compared to STARE, CHASE\_DB1 is more susceptible to bias as the images are all pairs from the same patient - this restricts the number of samples to 14. Due to this constraint and for the same reasons as with STARE, four-fold cross validation was used to preclude overlapping datasets between training and test time, this time grouping sets by patients. \footnote{https://blogs.kingston.ac.uk/retinal/chasedb1/}
\subsection{Preprocessing}  \label{sec:preproc}
The most common and effective method for correcting inconsistencies within an image dataset is by comparing the histogram of an image obtained to that of an ideal histogram describing the brightness, contrast and signal/noise ratio, and/or determination of image clarity by assessing morphological features \cite{fleming_2006}. Fundus images typically contain between 500$\times$500 to 2000$\times$2000 pixels, making training a classifier a memory and time consuming ordeal. Rather than processing the entire image, the images are randomly cropped and resized to 256$\times$256 pixels, flipped, rotated and/or enhanced to extend the dataset. 

\subsubsection{Continuous Pixel Space} \label{sec:pixel}
It has been shown that a continuous domain representation of pixel colour channels vastly improves memory efficiency during training \cite{kingma_2016}. This is primary due to dimensionality reduction from initial channel values to a distribution of [-0.5 to 0.5]; features are learned with densely packed gradients rather than needing to keep track of very sparse values associated with typical channel values \cite{salimans_2017}.

\subsubsection{Image enhancement} \label{sec:image_enhancement}
Local histogram enhancement methods greatly improve image quality and contrast, improving network performance during training and evaluation. Rather than sampling all pixels within an image once, histograms are generated for subsections of the image, each of which is normalized. One limitation for local methods is the risk of enhancing noise within the image. Contrast limited adaptive histogram equalization (CLAHE) is one method that overcomes this limitation. CLAHE limits the maximum pixel intensity peaks within a histogram, redistributing the values across all intensities prior histogram equalization \cite{szeliski_2010}. This is the contrast enhancement method used herein.
\subsection{Network Architecture} \label{sec:network}
 PixelBNN is a fully-residual autoencoder with gated residual streams, each initialized by differing convolutional filters. It is based on UNET\cite{ronneberger_2015}, PixelCNN \cite{oord_2016a} as well as various work on the use of skip connections and batch normalization within fully convolutional networks\cite{long_2015, drozdzal_2016, chen_2016,he_2016}. It differs from prior work in the layer architecture, use of gated filter streams and regularization by batch normalization joint with dropout during training. While nuanced, the network further differentiates from many state-of-the-art architectures in its use of Adam optimization, layer activation by CReLU and use of downsampling in place of other multi-resolution strategies. The network makes extensive use of CReLU to reduce feature redundancy and negative information loss that would otherwise be incurred with the use of rectified linear units (ReLU). CReLU models have been shown to consistently outperform ReLU models of equivalent size while reducing the number of parameters by half, leading to significant gains in performance \cite{shang_2016}.
 
The architecture was influenced by the human vision system: 
\begin{itemize}
	\item The use of two parallel input streams resembles bipolar cells in the retina, each stream possessing different yet potentially overlapping feature spaces initialized by different convolutional kernels.
	\item The layer structure is based on that of the lateral geniculate nucleus, visual cortices (V1, V2) and medial temporal gyrus, whereby each is represented by an encoder-decoder pair of gated resnet blocks. 
	\item Final classification is executed by a convolutional layer which concatenates the outputs of the last gated resnet block, as the inferotemporal cortex is believed to do.
\end{itemize}
More detail on this subject is covered in prior work by the authors\cite{dlra}.
\begin{figure}[!bp]
	\centering
	\includegraphics[width=1\linewidth]{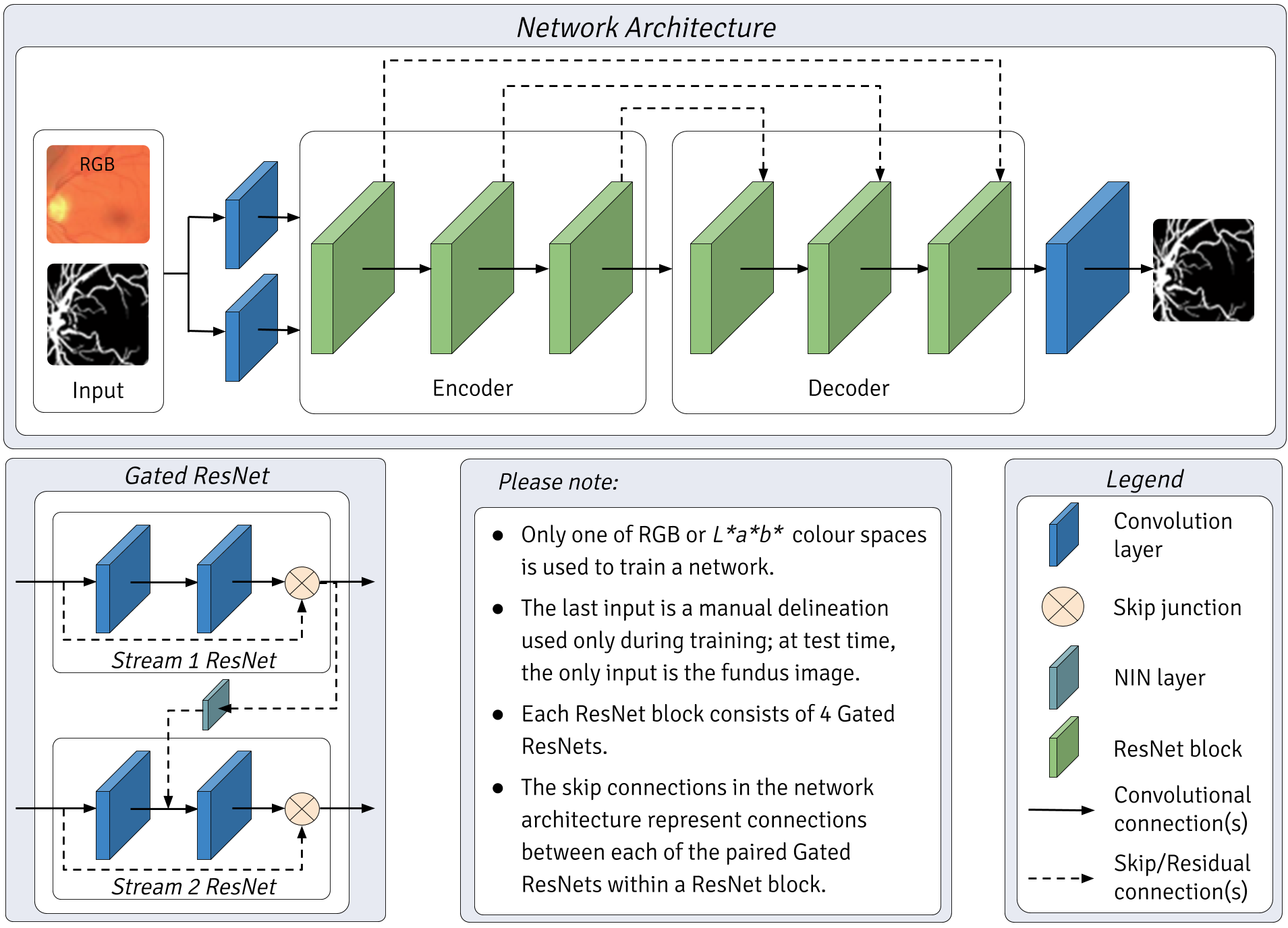}
	\caption{Processed image patches are passed through two convolution layers with different filters to create parallel input streams for the encoder. Downsampling occurs between each ResNet block in the encoder and upsampling in the decoder. The output is a vessel mask of equal size to the input. }
	\label{fig:pixelbnn}
\end{figure}

 \subsubsection{Downsampling without Information Loss} \label{sec:downsampling}
A popular method for facilitating multi-resolution generalizability with fully convolutional networks is the use of dilated convolutions within the model \cite{oord_2016a, kalchbrenner_2016a}. Dilated convolutions can be computational expensive, as they continuously increase in size through the utilization of zero padding to prevent information loss. Downsampling is another a family of methods that sample features during strided convolution at one or more intermediate stages of a FCN, later fusing the samples during upsampling \cite{long_2015} and/or multi-level classifiers \cite{chen_2016}. Such methods take advantage of striding to achieve similar processing improvements as dilated convolutions with increased computational efficiency, albeit with a loss in information. Variations in downsampling methods aim to compensate for this loss of information. 

\subsubsection{Proposed Method}  \label{sec:method}
Figure \ref{fig:pixelbnn} illustrates the architecture of the proposed method. PixelBNN utilizes downsampling with a stride of 2, as well as long and short skip connections, resembling PixelCNN++ \cite{salimans_2017}. Implementing both long and short skip connections has been shown to prevent information loss and increase convergence speed \cite{drozdzal_2016}, while mitigating losses in performance \cite{szegedy_2016}. The method differs from PixelCNN++ in three ways. First, feature maps are implemented as with UNET\cite{ronneberger_2015} with a starting value of 16, doubling at each downsampling. Second, in the use of batch normalization after each downsampling and before dropout, rather than dropout alone. Third, it differs in its use of paired convolution layers on on continuous pixel space RGB images. Each gated ResNet block consists of 4 gated ResNets as shown in Figure \ref{fig:pixelbnn}. Each ResNet is made up of convolution layers with kernel size 3 and stride of 1. Stream 1 ResNet is gated with Stream 2 by a network in network (NIN) layer, which is a 1x1 convolutional layer like those found in Inception models\cite{szegedy_2016}. 
\subsection{Platform}  \label{sec:platform}
Training and testing of the proposed method was done using a computer with an Intel(R) Core(TM) i7-5820K CPU with 3.30GHz of processing power, 32 GB of RAM and a GM200 GeForce GTX TITAN X graphics card equivalent to 3072 CUDA cores. On this platform, it took roughly 14 hours to train the network. At test time, the network processed a single image in 0.0466 seconds using the same system. In this study, Tensorflow\cite{tensorflow} and other python scientific, imaging and graphing libraries were used to evaluate the results. 
\section{Experimental Design} \label{sec:exp}
This paper presents PixelBNN, a novel network architecture for multi-resolution image segmentation and feature extraction based on PixelCNN. This is the first time this family of dense fully connected convolutional networks have been applied to fundus images. The specific task of retinal vessel segmentation was chosen due to the availability of different datasets that together provide ample variances for cross-validation, training efficiency, model performance and robustness. Architectural elements of the network have been thoroughly evaluated in the literature, as mentioned in Section \ref{sec:network}. An ablation study is beyond the scope of this paper and left for future work. Following the completion of the follow on study, the code will be made available here: \texttt{https://github.com/henryleopold/pixelbnn}

\subsection{Performance Indicators} \label{sec:kpis}
Model performance is evaluated using a set of key performance indicators (KPIs), which are calculated by comparing the network output against the first set of manual delineations as the ground truth on a per-pixel basis. The test dataset has a second set of manual delineations which are used to benchmark the results against a second human observer (the `2nd observer'). There are four potential classification outcomes for each pixel: true positive (TP), false positive (FP), true negative (TN) and false negative (FN). These outcomes are then used to derive KPIs, such as sensitivity (SN; also known as recall), specificity (SP), accuracy (Acc) and the receiver operating characteristic (ROC), which can be a function of SN and SP, true positive rate (TPR) and false positive rate (FPR) or other similar KPI pairs.  SN and SP are two of the most important KPIs to consider when developing a classification system as they are both representations of the ``truth condition'' and are thereby a far better performance measure than Acc. In an ideal system, both SN and SP will be 100\%, however this is rarely the case in real life. The area under a ROC curve (AUC) as well as Cohen's kappa coefficient ($\kappa$) are two common approaches for measuring network performance. $\kappa$ is measured using the probability ($n_{ki}$) of an observer ($i$) predicting a category ($k$) for a number of items ($N$) and provides a measure of agreement between observers - in this case, the network's prediction and the ground truth\cite{kappa}. 
\begin{table}[!bp]
	\centering
	\caption{Key Performance Indicators.}
	\label{tab:kpi}
	\renewcommand{\arraystretch}{2}
	\rowcolors{2}{gray!8}{white}
	\resizebox*{\columnwidth}{!}{%
		\begin{tabular}{m{0.3\linewidth} m{0.35\linewidth} M{0.35\linewidth}}
			\toprule
			\rule[-1ex]{0pt}{0ex}\textbf{KPI} &\textbf{Description} &\textbf{Value} \\
			\midrule
			\rule[0ex]{0pt}{3ex}True Positive Rate (TPR) & Probability of detection & \(\displaystyle \frac{TP}{vessel\ pixel\ count}\) \\[1.5ex]
			\rule[-1ex]{0pt}{3ex}False Positive Rate (FPR) & Probability of false detection &\rule[-1ex]{0pt}{5ex} \(\displaystyle \frac{FP}{non\-vessel\ pixel\ count}\) \\[1.5ex]
			\rule[-1ex]{0pt}{3ex}Accuracy (Acc) &The frequency a pixel is properly classified  &\(\displaystyle \frac{TP+TN}{total\ pixel\ count}\) \\
			\rule[-1ex]{0pt}{3ex}Sensitivity aka Recall (SN) &The proportion of true positive results detected by the classifier &$ TPR $ or \(\displaystyle \frac{TP}{TP+FN}\)  \\
			\rule[-1ex]{0pt}{3ex}Precision (Pr) &Proportion of positive samples properly classified & \(\displaystyle \frac{TP}{TP+FP}\) \\
			\rule[-1ex]{0pt}{3ex}Specificity (SP) &The proportion of negative samples properly classified & $1-FPR $ or \(\displaystyle \frac{TN}{TN+FP}\) \\
			\rule[-1ex]{0pt}{3ex}Kappa Coefficient ($\kappa$) & Agreement between two observers &\rule[-1ex]{0pt}{5ex} \(\displaystyle \frac{Acc-Acc_{prob}}{1-Acc_{prob}}\) \\
			\rule[-1ex]{0pt}{1ex}Probability of Agreement ($Acc_{prob}$ ) & Probability each observer $n_{ki}$ selects a category $k$ for $N$ items & \rule[-1ex]{0pt}{2ex}\(\displaystyle \frac{1}{N^{2}}\sum_{k}n_{k1}n_{k2}\) \\
			\rule[-1ex]{0pt}{2ex}G-mean (G) & Balance measure of SN and SP&  \(\displaystyle \sqrt{SN*SP}\) \\
			\rule[-1ex]{0pt}{2ex}Matthews Correlation Coefficient (MCC)&Measure from -1 to 1 for agreement between manual and predicted binary segmentations&$\aligned
				\ &\frac{(\sfrac{TP}{N})-S\times\,P}{\sqrt{P\times\,S\times\,(1-S)\times\,(1-P)}}\\[1ex]
				\ &\quad N=TP+FP+TN+FN\\[1ex]
				\ &\quad S=TP+FN\times\,N\\[1ex]
				\ &\quad P=TP+FP\times\,N
			\endaligned$ \\ 
			\rule[-1ex]{0pt}{3ex}F1 Score (F1) & Harmonic mean of precision and recall &  \(\displaystyle \frac{2*TP}{2TP + FP + FN}\)or\(\displaystyle \frac{2*Pr*SN}{Pr+SN}\)\\
			\bottomrule
		\end{tabular}%
		}
\end{table}

The Matthews Correlation Coefficient (MCC), the F1-score (F1) and the G-mean (G) performance metrics were used to better assess the resulting fundus label masks. These particular metrics are well suited for cases with imbalanced class ratios, as with the abundance of non-vessel pixels comparative to a low number of vessel pixels in this binary segmentation task. MCC has been used to assess vessel segmentation performance in several cases, and its value is a range from -1 to +1, respectively indicating total disagreement or alignment between the ground truth and prediction \cite{orlando_2017}. Precision (Pr) is the proportion on positive samples properly classified and is often measured against SN in a \textit{precision-recall curve}, similar to ROC. F-scores are harmonic means of Pr and SN and may incorporate weightings to adjust for class imbalances. This work uses the F1-score with a range from 0 to 1, where 1 signifies perfect segmentation of the positive class. G-mean calculates the geometric mean between SN and SP\cite{he_2009}.
The KPIs are defined in Table \ref{tab:kpi}.
 
\subsection{Training Details}  \label{sec:training_details}
\begin{table} [!bp] 
	\caption{Dataset Statistics} 
	\label{tab:dataset_stats}
	\renewcommand{\arraystretch}{1.1}
	\centering
	\resizebox{\columnwidth}{!}{%
		\begin{tabular}{l *{3}{M{0.3\linewidth}}}
			\toprule
			\textbf{\textit{Datasets}}&\textbf{DRIVE}&\textbf{STARE}&\textbf{CHASE\_DB1}\\\midrule
			\rule[-1ex]{0pt}{3ex} Image dimensions&565$\times$584&700$\times$605&1280$\times$960\\
			\rowcolor{gray!5}\rule[0ex]{0pt}{3ex} Colour Channels&RGB&RGB&RGB\\
			\rule[0ex]{0pt}{3ex} Total Images&40&20&28\\
			\rowcolor{gray!5}\rule[0ex]{0pt}{3ex} Source Grouping &20 train \& 20 test&-&14 Patients (2 images in each)\\
			\midrule
			 \multicolumn{4}{ l }{\textbf{\textit{Method Summary}}}\\\midrule
			\rule[-2ex]{0pt}{5ex} Train-Test Schedule &\makecell{One-off on 20 train\\Test on the other 20}&\makecell{4-fold cross-validation\\over 20 images}&\makecell{4-fold cross-validation\\over 14 patients}\\
			\rowcolor{gray!5}\rule[0ex]{0pt}{3ex} Information Loss&5.0348&6.4621&18.7500\\
		\bottomrule
		\end{tabular}%
	}
 \end{table}
For each dataset, the network parameters were randomly reinitialized using the Xavier algorithm\cite{xavier}. Table \ref{tab:dataset_stats} summarizes the three data sets as well as the test-train data distribution and approximate information loss incurred during preprocessing. Pre-training was never conducted and so the network was trained from scratch for each dataset; in the case of STARE and CHASE\_DB1, one set of parameters was trained from scratch for each fold. Images were reduced in size to alleviate the computational burden of the training task rather than using the original image to train the network. To ensure each dataset was evaluated equivalently, image size was first normalized to 256$\times$256 before undergoing dataset augmentation. This step is the cause for the majority of information loss relative to the original images and other methods compared herein which extract patches rather than resize the original fundus images.

The images were randomly cropped between 216 to 256 pixels along each axis and resized to 256$\times$256. They were then randomly flipped both horizontally and vertically before being rotated at zero, 90\textdegree \ or 180\textdegree. The brightness and contrast of each patch was randomly shifted to further increase network robustness. PixelBNN learns to generate vessel label masks from fundus images in batches of 3 for 100,000 iterations utilizing Adam optimization with an initial learning rate of $1e^{-5}$ and decay rate of 0.94 every 20,000 iterations. Batch normalization was conducted with an initial $\epsilon$ of $1e^{-5}$ and decay rate of 0.9 before the application of dropout regularization \cite{dropout} with a keep probability of 0.6. It required approximately 11 hours to complete training for DRIVE and the same for each fold during cross validation.
\section{Results}  \label{sec:results}
The output of PixelBNN is a binary label mask, predicting vessel and non-vessel pixels thereby segmenting the original image. Each dataset contains a two experts' manual delineations; the first was used as the ground truth for training the model and the second was used for evaluating the network's performance against a secondary human observer. Independently, each dataset was used to train a separate model from scratch resulting in three sets of model parameters. 

\subsection{Performance Comparison} \label{sec:comparison}
\newlength{\driveseg}
\newcommand{\rowname}[1]
{\rotatebox{90}{\makebox[\driveseg][c]{\textbf{#1}}}}
\renewcommand{\thesubfigure}{\alph{subfigure}}
\newcommand{\mycaption}[1]
{\refstepcounter{subfigure}\textbf{(\thesubfigure) }{\ignorespaces #1}}
\begin{figure}[!tp]
	\settoheight{\driveseg}{\includegraphics[width=.2\linewidth]{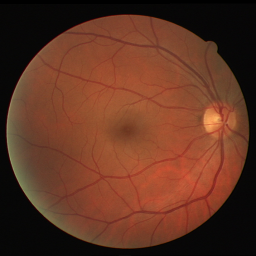}}%
	\centering
	\begin{tabular}{@{}c@{ }c@{ }c@{ }c@{ }c@{ }c@{}}
		&\textbf{Image}& \textbf{DRIVE} & \textbf{STARE}  & \textbf{CHASE\_DB1}& \textbf{Ground Truth}\\
		\rowname{Best}&
		\includegraphics[width=.19\linewidth]{imagesets/drive_drive/19.png}&
		\includegraphics[width=.19\linewidth]{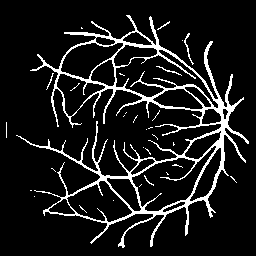}&
		\includegraphics[width=.19\linewidth]{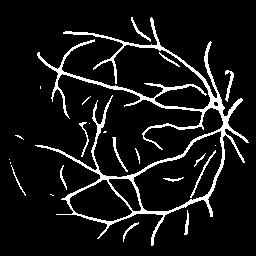}&
		\includegraphics[width=.19\linewidth]{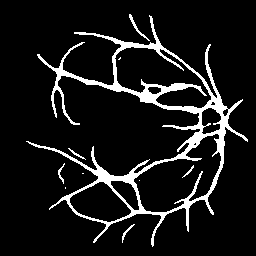}&
		\includegraphics[width=.19\linewidth]{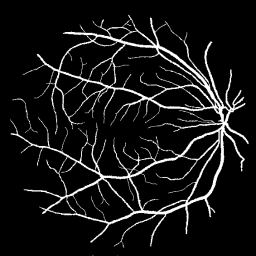}\\[-1ex]
		\rowname{Worst}&
		\includegraphics[width=.19\linewidth]{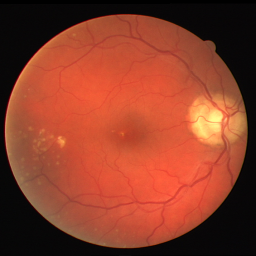}&
		\includegraphics[width=.19\linewidth]{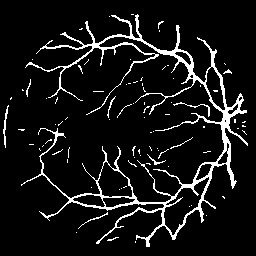}&
		\includegraphics[width=.19\linewidth]{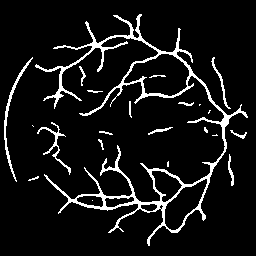}&
		\includegraphics[width=.19\linewidth]{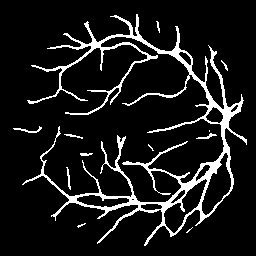}&
		\includegraphics[width=.19\linewidth]{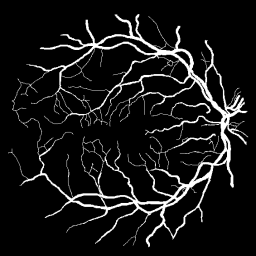}\\
	\end{tabular}
	\caption{Network predictions on the DRIVE dataset. The top row shows the image, segmentation masks and ground truth for the image that scored best when DRIVE was used to train and test the model; the bottom row shows the worst. For comparison, the cross-validation results from training the model with STARE and CHASE\_DB1 are shown.}
	\label{fig:drive_segs}
\end{figure}

\newlength{\stareseg}
\renewcommand{\rowname}[1]
{\rotatebox{90}{\makebox[\stareseg][c]{\textbf{#1}}}}
\renewcommand{\thesubfigure}{\alph{subfigure}}
\renewcommand{\mycaption}[1]
{\refstepcounter{subfigure}\textbf{(\thesubfigure) }{\ignorespaces #1}}
\begin{figure}[!bp]
	\settoheight{\stareseg}{\includegraphics[width=.2\linewidth]{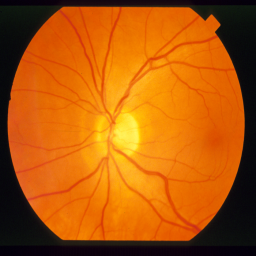}}%
	\centering
	\begin{tabular}{@{}c@{ }c@{ }c@{ }c@{ }c@{ }c@{}}
		&\textbf{Image}& \textbf{STARE} & \textbf{DRIVE}  & \textbf{CHASE\_DB1}& \textbf{Ground Truth}\\
		\rowname{Best}&
		\includegraphics[width=.19\linewidth]{imagesets/stare_stare/0163.png}&
		\includegraphics[width=.19\linewidth]{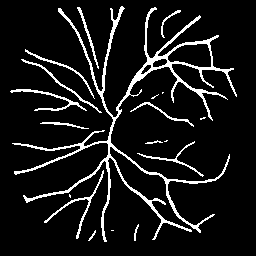}&
		\includegraphics[width=.19\linewidth]{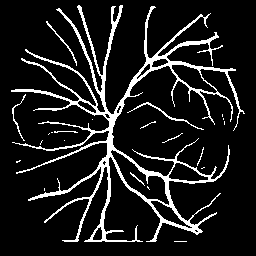}&
		\includegraphics[width=.19\linewidth]{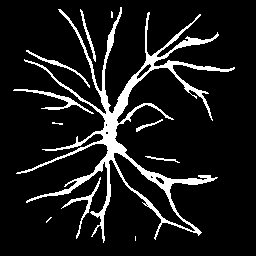}&
		\includegraphics[width=.19\linewidth]{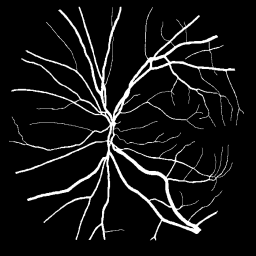}\\[-1ex]
		\rowname{Worst}&
		\includegraphics[width=.19\linewidth]{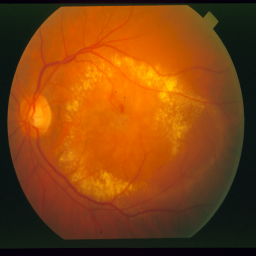}&
		\includegraphics[width=.19\linewidth]{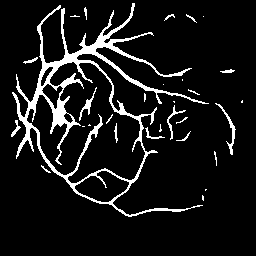}&
		\includegraphics[width=.19\linewidth]{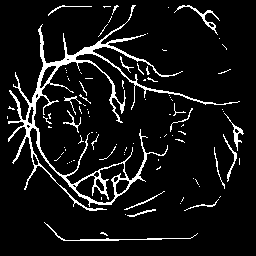}&
		\includegraphics[width=.19\linewidth]{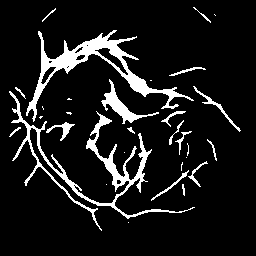}&
		\includegraphics[width=.19\linewidth]{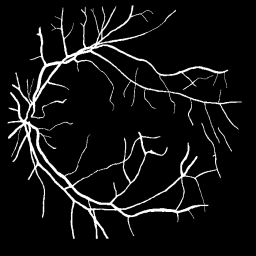}\\
	\end{tabular}
	\caption{Network predictions on the STARE dataset. The top row shows the image, segmentation masks and ground truth for the image that scored best when STARE was used to train and test the model; the bottom row shows the worst. For comparison, the cross-validation results from training the model with DRIVE and CHASE\_DB1 are shown.} 
	\label{fig:stare_segs}
\end{figure}

\newlength{\tempdima}
\renewcommand{\rowname}[1]
{\rotatebox{90}{\makebox[\tempdima][c]{\textbf{#1}}}}
\renewcommand{\thesubfigure}{\alph{subfigure}}
\renewcommand{\mycaption}[1]
{\refstepcounter{subfigure}\textbf{(\thesubfigure) }{\ignorespaces #1}}
\begin{figure}[!bp]
	\settoheight{\tempdima}{\includegraphics[width=.2\linewidth]{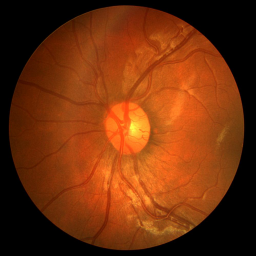}}%
	\centering
	\begin{tabular}{@{}c@{ }c@{ }c@{ }c@{ }c@{ }c@{}}
		&\textbf{Image}& \textbf{CHASE\_DB1} & \textbf{STARE}  & \textbf{DRIVE}& \textbf{Ground Truth}\\
		\rowname{Best}&
		\includegraphics[width=.19\linewidth]{imagesets/chase_chase/02L.png}&
		\includegraphics[width=.19\linewidth]{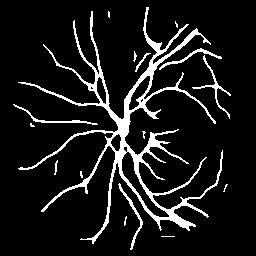}&
		\includegraphics[width=.19\linewidth]{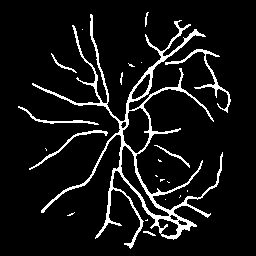}&
		\includegraphics[width=.19\linewidth]{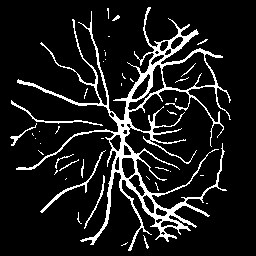}&
		\includegraphics[width=.19\linewidth]{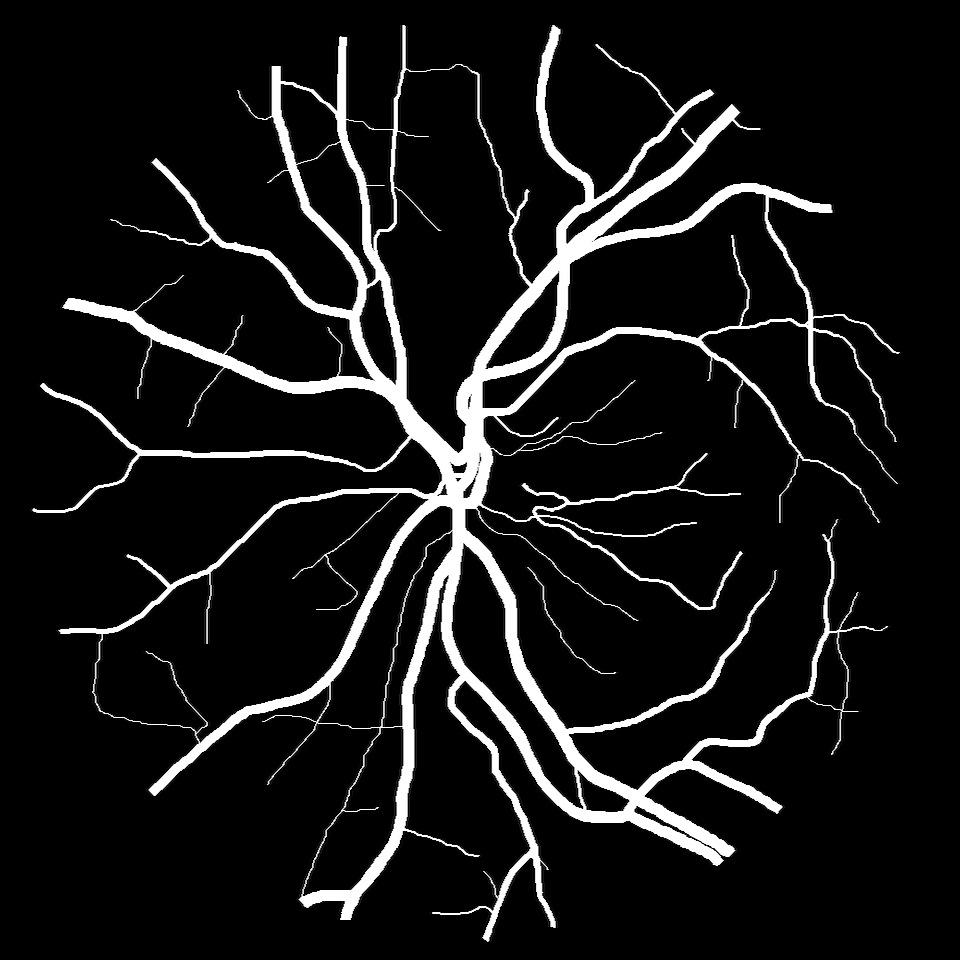}\\[-1ex]
		\rowname{Worst}&
		\includegraphics[width=.19\linewidth]{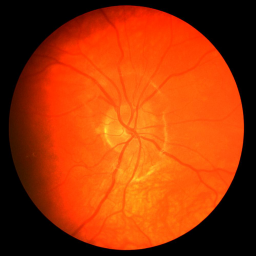}&
		\includegraphics[width=.19\linewidth]{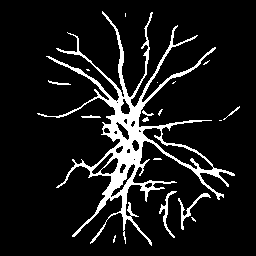}&
		\includegraphics[width=.19\linewidth]{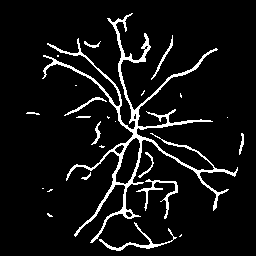}&
		\includegraphics[width=.19\linewidth]{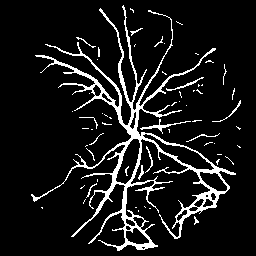}&
		\includegraphics[width=.19\linewidth]{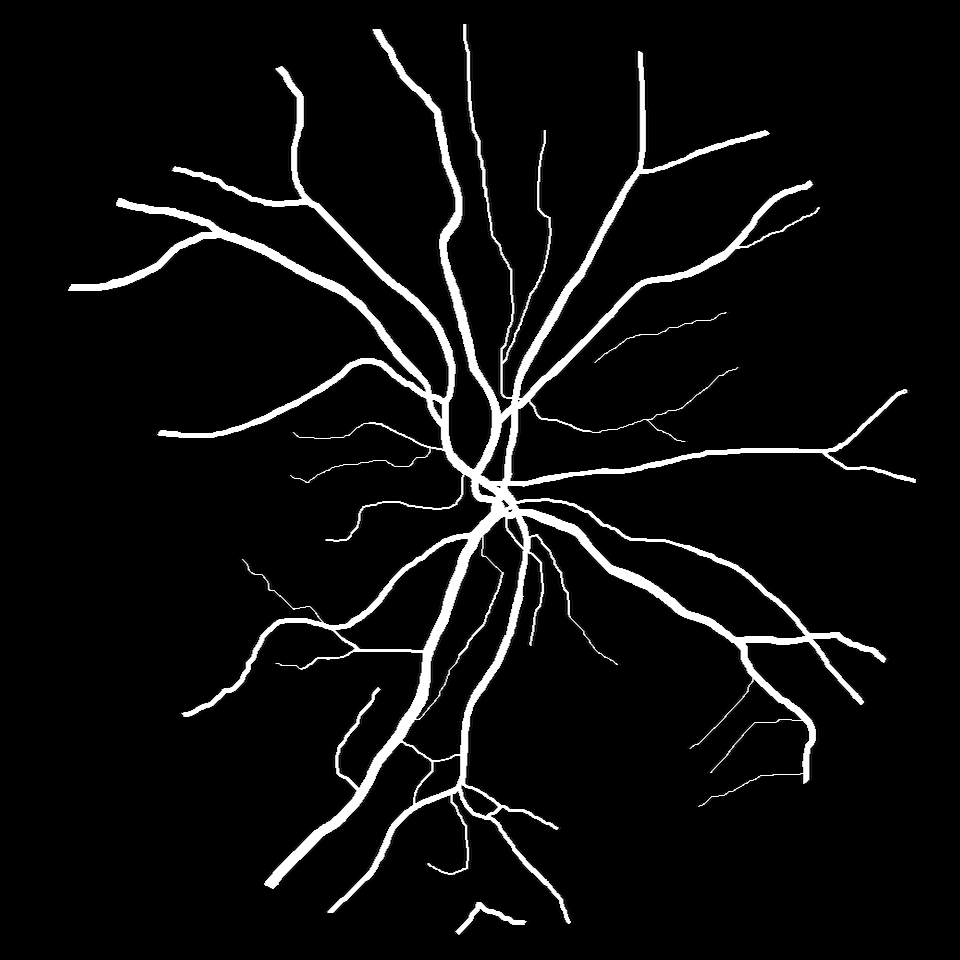}\\
	\end{tabular}
	\caption{Network predictions on the CHASE\_DB1 dataset. The top row shows the image, segmentation masks and ground truth for the image that scored best when CHASE\_DB1 was used to train and test the model; the bottom row shows the worst. For comparison, the cross-validation results from training the model with STARE and DRIVE are shown.}
	\label{fig:chase_segs}
\end{figure}

The results were compared with those of other state-of-the-art methods for vessel segmentation with published results for at least one of the DRIVE, STARE or CHASE\_DB1 datasets. The results for the model trained and tested on DRIVE are shown in Table \ref{tab:drive_comp}, STARE results are shown in Table \ref{tab:stare_comp} and CHASE\_DB1 results are in Table \ref{tab:chase_comp}. Cross-testing was conducted using each of these sets to measure the performance of the network against each other datasets' test images. The results from cross-testing are summarized in Table \ref{tab:cross_comp}. Most of the articles report SN and SP, relying on Acc and AUC to validate performance, whereas $\kappa$, MCC and F1-scores have been sparsely applied until recently. Regardless of other KPIs, most recent works report SN and SP from which the G-mean was calculated. Herein, the G-mean is considered to be a truer performance indicator than SN, SP and Pr. Further, the main KPIs used to evaluate model performance are F1-score, G-mean and MCC. For completeness, SN, SP, Pr, Acc, AUC and $\kappa$ are also tabulated. Table \ref{tab:computation_time} compares the computation time for training the network and evaluating test images with the methods that share the same GPU. 

\begin{table} [!tbp]
	\caption{Performance comparison for models trained and tested with DRIVE.} 
	\label{tab:drive_comp}
	\renewcommand{\arraystretch}{0.9}
	\centering
	\resizebox{\columnwidth}{!}{%
		\begin{tabular}{l *{9}{M{0.095\linewidth}}}
			\toprule
			\rule[-1ex]{0pt}{2ex} \textbf{Methods}&\textbf{SN}&\textbf{SP}&\textbf{Pr}&\textbf{Acc}&\textbf{AUC}&\textbf{kappa}&\textbf{G}&\textbf{MCC}&\textbf{F1}\\
			\midrule
			\rule[-1ex]{0pt}{2ex}  Human (2nd Observer)&0.7760&0.9730&0.8066&0.9472&-&0.7581&0.8689&0.7601&0.7881\\\midrule 
			\multicolumn{10}{l}{\,\textit{\textbf{Unsupervised Methods}} } \\\midrule
			\rule[-1ex]{0pt}{2ex} Lam et al. \cite{lam_2010}&-&-&-&0.9472&0.9614&-&-&-&-\\
			\rule[-1ex]{0pt}{2ex} Azzopardi et al. \cite{azzopardi_2015}&0.7655&0.9704&-&0.9442&0.9614&-&0.8619&0.7475&-\\
			\rule[-1ex]{0pt}{2ex} Kov\'{a}cs and Hajdu \cite{kovacs_2016}&0.7270&0.9877&-&0.9494&-&-&0.8474&-&- \\
			\rule[-1ex]{0pt}{2ex} Zhang et al. \cite{zhang_2016}&0.7743&0.9725&-&0.9476&0.9636&-&0.8678&-&-\\
			\rule[-1ex]{0pt}{2ex} Roychowdhury et al. \cite{roychowdhury_2015}&0.7395$\pm$ 0.062&0.9782$\pm$ 0.0073&-&0.9494$\pm$ 0.005&0.9672&-&-&-&- \\
			\rule[-1ex]{0pt}{2ex} Niemeijer et al. \cite{niemeijer_2004}&0.6793$\pm$ 0.0699&0.9801$\pm$ 0.0085&-&0.9416$\pm$ 0.0065&9294$\pm$ 0.0152&0.7145&-&-&-\\
			\midrule \multicolumn{10}{l}{\,\textit{\textbf{Supervised Methods}} } \\\midrule
			\rule[-1ex]{0pt}{2ex} Soares et al. \cite{soares_2006}&0.7332&0.9782&-&0.9461$\pm$ 0.0058&0.9614&0.7285&0.8469&-&-\\
			\rule[-1ex]{0pt}{2ex} Ricci and Perfetti \cite{ricci_2007}&-&-&-&0.9595&0.9633&-&-&-&-\\
			\rule[-1ex]{0pt}{2ex} Marin et al. \cite{marin_2011}&0.7067&0.9801&-&0.9452&0.9588&-&0.8322&-&-\\
			\rule[-1ex]{0pt}{2ex} Lupascu et al. \cite{lupascu_2010}&-&-&-&0.9597$\pm$ 0.0054&0.9561&0.7200&0.8151&-&-\\
			\rule[-1ex]{0pt}{2ex} Fraz et al. \cite{fraz_2011}&0.7152&0.9768&0.8205&0.9430&-&-&0.8358&0.7333&0.7642\\
			\rule[-1ex]{0pt}{2ex} Fraz et al. \cite{fraz_2012b}&0.7406&0.9807&-&0.9480&0.9747&-&0.8522&-&-\\
			\rule[-1ex]{0pt}{2ex} Fraz et al. \cite{fraz_2013}&0.7302&0.9742&0.8112&0.9422&-&-&0.8434&0.7359&0.7686\\
			\rule[-1ex]{0pt}{2ex} Vega et al. \cite{vega_2015}&0.7444&0.9600&-&0.9412&-&-&0.8454&0.6617&0.6884\\ 
			\rule[-1ex]{0pt}{2ex} Li et al. \cite{li_2016}&0.7569&0.9816&-&0.9527&0.9738&-&0.8620&-&-\\
			\rule[-1ex]{0pt}{2ex} Liskowski et al. \cite{liskowski_2016}&0.7811&0.9807&-&0.9535&0.9790&0.7910&0.8752&-&-\\
			\rule[-1ex]{0pt}{2ex} Leopold et al. \cite{leopold_2017b}&0.6823&0.9801&-&0.9419&0.9707&-&0.8178&-&-\\
			\rule[-1ex]{0pt}{2ex} Leopold et al. \cite{leopold_2017a}&0.7800&0.9727&-&0.9478&0.9689&-&0.8710&-&-\\
			\rule[-1ex]{0pt}{2ex} Orlando et al. \cite{orlando_2017}&0.7897&0.9684&0.7854&-&-&-&0.8741&0.7556&0.7857\\
			\rule[-1ex]{0pt}{2ex} Mo et al. \cite{mo_2017}&0.7779$\pm$ 0.0849&0.9780$\pm$ 0.0091&-&0.9521$\pm$ 0.0057&0.9782$\pm$ 0.0059&0.7759$\pm$ 0.0329&0.8722$\pm$ 0.0278&-&-\\
			\midrule
			\rule[-1ex]{0pt}{2ex} \emph{PixelBNN}&0.6963$\pm$ 0.0489&0.9573$\pm$ 0.0089&0.7770$\pm$ 0.0458&0.9106$\pm$ 0.0121&0.8268$\pm$ 0.0247&0.6795$\pm$ 0.0414&0.8159$\pm$ 0.0286&0.6820$\pm$ 0.0399&0.7328$\pm$ 0.0335\\	
		\bottomrule
		\end{tabular}%
	}
 \end{table}

\begin{table} [!tbp]
	\caption{Performance comparison for models trained and tested with STARE.} 
	\label{tab:stare_comp}
	\renewcommand{\arraystretch}{0.9}
	\centering
	\resizebox{\textwidth}{!}{%
		\begin{tabular}{l *{9}{M{0.099\linewidth}} }
			\toprule
			\rule[-1ex]{0pt}{2ex} \textbf{Methods}& \textbf{SN}&\textbf{SP}&\textbf{Pr}&\textbf{Acc}&\textbf{AUC}&\textbf{kappa}&\textbf{G}&\textbf{MCC}&\textbf{F1}\\
			\midrule
			\rule[-1ex]{0pt}{2ex}  Human (2nd Observer)&0.8951&0.9387&0.6424&0.9353&-&0.7046&0.9166&0.7225&0.7401\\
			\midrule \multicolumn{10}{l}{\,\textit{\textbf{Unsupervised Methods}} } \\\midrule
				\rule[-1ex]{0pt}{2ex} Lam et al. \cite{lam_2010}&-&-&-&0.9567&0.9739&-&-&-&-\\
			\rule[-1ex]{0pt}{2ex} Azzopardi et al. \cite{azzopardi_2015}&0.7716&0.9701&-&0.9497&0.9563&-&0.8652&0.7335&-\\
			\rule[-1ex]{0pt}{2ex} Kov\'{a}cs and Hajdu \cite{kovacs_2016}&0.7665&0.9879&-&-&0.9711&-&0.8702&-&-\\
			\rule[-1ex]{0pt}{2ex} Zhang et al. \cite{zhang_2016}&0.7791&0.9758&-&0.9554&0.9748&-&0.8719&-&-\\
			\rule[-1ex]{0pt}{2ex} Roychowdhury et al. \cite{roychowdhury_2015}&0.7317$\pm$ 0.053&0.9842$\pm$ 0.0069&-&0.9560$\pm$ 0.0095&0.9673&-&0.8486$\pm$ 0.0178&-&-\\
			\midrule \multicolumn{10}{l}{\,\textit{\textbf{Supervised Methods}} } \\\midrule
			\rule[-1ex]{0pt}{2ex} Soares et al. \cite{soares_2006}&0.7207&0.9747&-&0.9479&0.9671&-&0.8381&-&- \\
			\rule[-1ex]{0pt}{2ex} Ricci et al. \cite{ricci_2007}&-&-&-&0.9584&0.9602&-&-&-&- \\
			\rule[-1ex]{0pt}{2ex} Marin et al. \cite{marin_2011}&0.6944&0.9819&-&0.9526&0.9769&-&0.8257&-&-\\
			\rule[-1ex]{0pt}{2ex} Fraz et al. \cite{fraz_2011}&0.7409&0.9665&0.7363&0.9437&-&-&0.8462&0.7003&0.7386\\
			
			\rule[-1ex]{0pt}{2ex} Fraz et al. \cite{fraz_2012b}&0.7548&0.9763&-&0.9534&0.9768&-&0.8584&-&-\\
			\rule[-1ex]{0pt}{2ex} Fraz et al. \cite{fraz_2013}&0.7318&0.9660&0.7294&0.9423&-&-&0.8408&0.6908&0.7306\\
			\rule[-1ex]{0pt}{2ex} Vega et al. \cite{vega_2015}&0.7019&0.9671&-&0.9483&-&-&0.8239&0.5927&0.6082\\ 
			\rule[-1ex]{0pt}{2ex} Li et al. \cite{li_2016}&0.7726&0.9844&-&0.9628&0.9879&-&0.8721&-&-\\
			\rule[-1ex]{0pt}{2ex} Liskowski et al. \cite{liskowski_2016}&0.8554$\pm$ 0.0286&0.9862$\pm$ 0.0018&-&0.9729$\pm$ 0.0027&0.9928$\pm$ 0.0014&0.8507$\pm$ 0.0155&0.9185$\pm$ 0.0072&-&-\\
			\rule[-1ex]{0pt}{2ex} Mo et al. \cite{mo_2017}&0.8147$\pm$ 0.0387&0.9844$\pm$ 0.0034&-&0.9674$\pm$ 0.0058&0.9885$\pm$ 0.0035&0.8163$\pm$ 0.0310&0.8955$\pm$ 0.0115&-& -\\
			\rule[-1ex]{0pt}{2ex} Orlando et al. \cite{orlando_2017}&0.7680&0.9738&0.7740&-&-&-&0.8628&0.7417&0.7644\\
			\midrule
			\rule[-1ex]{0pt}{2ex} \emph{PixelBNN}&0.6433$\pm$ 0.0593&0.9472$\pm$ 0.0212&0.6637$\pm$ 0.1135&0.9045$\pm$ 0.0207&0.7952$\pm$ 0.0315&0.5918$\pm$ 0.0721&0.7797$\pm$ 0.0371&0.5960$\pm$ 0.0719&0.6465$\pm$ 0.0621\\
		\bottomrule
		\end{tabular}%
	}
 \end{table}

\begin{table} [!tbp]
	\caption{Performance comparison for models trained and tested with CHASE\_DB1.} 
	\label{tab:chase_comp}
	\renewcommand{\arraystretch}{0.9}
	\centering
	\resizebox{\textwidth}{!}{%
		\begin{tabular}{l *{9}{M{0.095\linewidth}} }
			\toprule
			\rule[-1ex]{0pt}{2ex} \textbf{Methods}&\textbf{SN}&\textbf{SP}&\textbf{Pr}&\textbf{Acc}&\textbf{AUC}&\textbf{kappa}&\textbf{G}&\textbf{MCC}&\textbf{F1}\\
			\midrule
			\rule[-1ex]{0pt}{2ex}  Human (2nd Observer)&0.7425&0.9793&0.8090&0.9560&-&0.7529&0.8527&0.7475&0.7686\\
	  		\midrule \multicolumn{10}{l}{\,\textit{\textbf{Unsupervised Methods}} } \\\midrule
			\rule[-1ex]{0pt}{2ex} Azzopardi et al. \cite{azzopardi_2015}&0.7585&0.9587&-&0.9387&0.9487&-&0.8527&0.6802&-\\
			\rule[-1ex]{0pt}{2ex} Zhang et al. \cite{zhang_2016}&0.7626&0.9661&-&0.9452&0.9606&-&0.8583&-&-\\
			\rule[-1ex]{0pt}{2ex} Roychowdhury et al. \cite{roychowdhury_2015}&0.7615$\pm$ 0.0516&0.9575$\pm$ 0.003&-&0.9467$\pm$ 0.0076&0.9623&-&0.8539$\pm$ 0.0124&-&-\\
			\midrule \multicolumn{10}{l}{\,\textit{\textbf{Supervised Methods}} } \\\midrule
			\rule[-1ex]{0pt}{2ex} Fraz et al. \cite{fraz_2012b}&0.7224&0.9711&-&0.9469&0.9712&-&0.8376&-&-\\
			\rule[-1ex]{0pt}{2ex} Li et al. \cite{li_2016}&0.7507&0.9793&-&0.9581&0.9716&-&0.8574&-&-\\
			\rule[-1ex]{0pt}{2ex} Liskowski et al. \cite{liskowski_2016}&0.7816$\pm$ 0.0178&0.9836$\pm$ 0.0022&-&0.9628$\pm$ 0.0020&0.9823$\pm$ 0.0016&0.7908$\pm$ 0.0111&0.8768$\pm$ 0.0063&-&-  \\
	   		\rule[-1ex]{0pt}{2ex} Mo et al. \cite{mo_2017}&0.7661 $\pm$ 0.0533&0.9816$\pm$ 0.0076&-&0.9599$\pm$ 0.0050&0.9812$\pm$ 0.0040&0.8672$\pm$ 0.0201&0.7689$\pm$ 0.0263&-& -\\
	 		\rule[-1ex]{0pt}{2ex} Orlando et al. \cite{orlando_2017}&0.7277&0.9712&0.7438&-&-&-&0.8403&0.7046&0.7332\\
			\midrule
			\rule[-1ex]{0pt}{2ex} \emph{PixelBNN}&0.8618$\pm$ 0.0232&0.8961$\pm$ 0.0150&0.3951$\pm$ 0.0603&0.8936$\pm$ 0.0138&0.878959$\pm$ 0.0138&0.4889$\pm$ 0.0609&0.8787$\pm$ 0.0140&0.5376$\pm$ 0.0491&0.5391$\pm$ 0.0587\\
			\bottomrule
		\end{tabular}%
	}
 \end{table}

\begin{table} [!tbp]
	\caption{Model performance measures from cross-training.} 
	\label{tab:cross_comp}
	\renewcommand{\arraystretch}{0.8}
	\centering
	\resizebox{\textwidth}{!}{%
		\begin{tabular}{c l*{9}{M{0.093\linewidth}} }
			\toprule
			\rule[-1ex]{0pt}{2ex}&\textbf{Methods}&\textbf{SN}&\textbf{SP}&\textbf{Pr}&\textbf{Acc}&\textbf{AUC}&\textbf{kappa}&\textbf{G}&\textbf{MCC}&\textbf{F1}\\
	  		\midrule\multicolumn{10}{l}{\quad\textit{Test images from: \textbf{DRIVE}} } \\\midrule
			\multirow{7}{*}{\makecell{\textit{Model}\\\textit{trained on:} \\ \textbf{STARE}}}&Soares et al. \cite{soares_2006}&-&-&-&0.9397&-&-&-&-&-\\
			\rule[-1ex]{0pt}{2ex}&Ricci et al. \cite{ricci_2007}&-&-&-&0.9266&-&-&-&-&-\\
			\rule[-1ex]{0pt}{2ex}&Marin et al. \cite{marin_2011}&-&-&-&0.9448&-&-&-&-&-\\
			\rule[-1ex]{0pt}{2ex}&Fraz et al. \cite{fraz_2012b}&0.7242&0.9792&-&0.9456&0.9697&-&0.8421&-&-\\
			\rule[-1ex]{0pt}{2ex}&Li et al. \cite{li_2016}&0.7273&0.9810&-&0.9486&0.9677&-&0.8447&-&-\\
			\rule[-1ex]{0pt}{2ex}&Liskowski et al. \cite{liskowski_2016}&-&-&-&0.9416&0.9605&-&-&-&-  \\
			\rule[-1ex]{0pt}{2ex}&Mo et al. \cite{mo_2017}&0.7412&0.9799&-&0.9492&0.9653&-&0.8522&-&-\\
			
			\rule[-1ex]{0pt}{2ex}&\textit{PixelBNN}&0.5110$\pm$ 0.0362&0.9533$\pm$ 0.0094&0.7087$\pm$ 0.0554&0.8748$\pm$ 0.0126&0.7322$\pm$ 0.0199&0.5193$\pm$ 0.0404&0.6974$\pm$ 0.0258&0.5309$\pm$ 0.0422&0.5907$\pm$ 0.0348\\
			\midrule
			
			\multirow{3}{*}{\makecell{\textit{Model}\\\textit{trained on:}\\\textbf{CHASE\_DB1}}}&Li et al. \cite{li_2016}&0.7307&0.9811&-&0.9484&0.9605&-&0.8467&-&-\\
			\rule[-1ex]{0pt}{2ex}&Mo et al. \cite{mo_2017}&0.7315&0.9778&-&0.9460&0.9650&-&0.8457&-&-\\
			
			\rule[-1ex]{0pt}{2ex}&\textit{PixelBNN} 
			&0.6222$\pm$ 0.0441&0.9355$\pm$ 0.0085&0.6785$\pm$ 0.0383&0.8796$\pm$ 0.0090&0.7788$\pm$ 0.0204&0.5742$\pm$ 0.0282&0.7622$\pm$ 0.0254&0.5768$\pm$ 0.0279&0.6463$\pm$ 0.0237 \\
			\midrule
	  		\multicolumn{10}{l}{\quad\textit{Test images from: \textbf{STARE}} } \\\midrule
			 \multirow{7}{*}{\makecell{\textit{Model}\\\textit{trained on:}\\\textbf{DRIVE}}}&Soares et al. \cite{soares_2006}&-&-&-&0.9327&-&-&-&-&-\\
			\rule[-1ex]{0pt}{2ex}&Ricci et al. \cite{ricci_2007}&-&-&-&0.9464&-&-&-&-&-\\
			\rule[-1ex]{0pt}{2ex}&Marin et al. \cite{marin_2011}&-&-&-&0.9528&-&-&-&-&-\\
			\rule[-1ex]{0pt}{2ex}&Fraz et al. \cite{fraz_2012b}&0.7010&0.9770&-&0.9493&0.9660&-&0.8276&-&-\\
			\rule[-1ex]{0pt}{2ex}&Li et al. \cite{li_2016}&0.7027&0.9828&-&0.9545&0.9671&-&0.8310&-&-\\
			\rule[-1ex]{0pt}{2ex}&Liskowski et al. \cite{liskowski_2016}&-&-&-&0.9505&0.9595&-&-&-&-  \\
			\rule[-1ex]{0pt}{2ex}&Mo et al. \cite{mo_2017}&0.7009&0.9843&-&0.9570&0.9751&-&0.8306&-&-\\
			
			\rule[-1ex]{0pt}{2ex}&\textit{PixelBNN}&0.7842$\pm$ 0.0552&0.9265$\pm$ 0.0196&0.6262$\pm$ 0.1143&0.9070$\pm$ 0.0181&0.8553$\pm$ 0.0323&0.6383$\pm$ 0.0942&0.8519$\pm$ 0.0343&0.6465$\pm$ 0.0873&0.6916$\pm$ 0.0868\\
			\midrule
			
	 		\multirow{3}{*}{\makecell{\textit{Model}\\\textit{trained on:} \\ \textbf{CHASE\_DB1}}}&Li et al. \cite{li_2016}&0.6944&0.9831&-&0.9536&0.9620&-&0.8262&-&-\\
	 		\rule[-1ex]{0pt}{2ex}&Mo et al. \cite{mo_2017}&0.7387&0.9787&-&0.9549&0.9781&-&0.8503&-&-\\
	 		
	 		\rule[-1ex]{0pt}{2ex}&\textit{PixelBNN} 
	 		&0.6973$\pm$ 0.0372&0.9062$\pm$ 0.0189&0.5447$\pm$ 0.0957&0.8771$\pm$ 0.0157&0.8017$\pm$ 0.0226&0.5353$\pm$ 0.0718&0.7941$\pm$ 0.0245&0.5441$\pm$ 0.0649&0.6057$\pm$ 0.0674\\
	 		\midrule
	 		
	 		\multicolumn{10}{l}{\quad\textit{Test images from: \textbf{CHASE\_DB1}} } \\\midrule
			 \multirow{3}{*}{\makecell{\textit{Model}\\\textit{trained on:} \\ \textbf{DRIVE}}}&Li et al. \cite{li_2016}&0.7118&0.9791&-&0.9429&0.9628&-&0.8348&-&-\\
			\rule[-1ex]{0pt}{2ex}&Mo et al. \cite{mo_2017}&0.7003&0.9750&-&0.9478&0.9671&-&0.8263&-&-\\
			
			\rule[-1ex]{0pt}{2ex}&\textit{PixelBNN}&0.9038$\pm$ 0.0196&0.8891$\pm$ 0.0089&0.3886$\pm$ 0.0504&0.8901$\pm$ 0.0088&0.8964$\pm$ 0.0116&0.4906$\pm$ 0.0516&0.8963$\pm$ 0.0116&0.5480$\pm$ 0.0413&0.5416$\pm$ 0.0513\\
			\midrule
			
			\multirow{4}{*}{\makecell{\textit{Model}\\\textit{trained on:} \\ \textbf{STARE}}}&Fraz et al. \cite{fraz_2012b}&0.7103&0.9665&-&0.9415&0.9565&-&0.8286&-&-\\
			\rule[-1ex]{0pt}{2ex}&Li et al. \cite{li_2016}&0.7240&0.9768&-&0.9417&0.9553&-&0.8410&-&-\\
			\rule[-1ex]{0pt}{2ex}&Mo et al. \cite{mo_2017}&0.7032&0.9794&-&0.9515&0.9690&-&0.8299&-&-\\
			
			\rule[-1ex]{0pt}{2ex}&\textit{PixelBNN}&0.7525$\pm$ 0.0233&0.9302$\pm$ 0.0066&0.4619$\pm$ 0.0570&0.9173$\pm$ 0.0059&0.8413$\pm$ 0.0132&0.5266$\pm$ 0.0482&0.8365$\pm$ 0.0143&0.5475$\pm$ 0.0412&0.5688$\pm$ 0.0475\\
			\bottomrule
		\end{tabular}%
	}
 \end{table}

\begin{table} [!tbp]
	\caption{Computation time for different networks using an NVIDIA Titan X.} 
	\label{tab:computation_time}
	\renewcommand{\arraystretch}{0.2}
	\rowcolors{2}{gray!8}{white}
	\centering
	\resizebox{0.85\columnwidth}{!}{%
		\begin{tabular}{m{0.2\linewidth} m{0.4\linewidth} *{2}{M{0.15\linewidth}}}
			\toprule
			\textbf{Method}&\textbf{Description}&\textbf{Training time} (\textit{s/iteration})&\textbf{Test time} (\textit{s/image})\\
			\midrule
			\rule[-1ex]{0pt}{3.5ex} Liskowski et al.\cite{liskowski_2016}&Repurposed MNIST LeNet&0.96&92\\
			\rule[-1ex]{0pt}{3.5ex} Mo et al. \cite{mo_2017}&Pre-trained Multi-classifier&\textit{N/A}&0.4\\
			\rule[-1ex]{0pt}{3.5ex} PixelBNN&\textit{Proposed Method}&0.52&0.0466\\ 
			\bottomrule
		\end{tabular}%
	}
\end{table}

The model's performance varied between datasets, outperforming other methods in a subset of cross-testing tasks for which there were few published baselines. At face value, the model appears to underperform the state-of-the-art, however the information lost when resizing the images during preprocessing is quite severe. Figure \ref{fig:drive_segs}, Figure \ref{fig:stare_segs} and Figure \ref{fig:chase_segs} show the best and worst scoring same-set images, ground truth and resulting predictions for testing and cross-testing that image with DRIVE, STARE and CHASE\_DB1 respectively. Overall, the predictions reveal that losses in performance are largely the result of fine-vessels being missed as well as anomalous pathologies. Interestingly, PixelBNN performed better on STARE and CHASE\_DB1 when the model was trained with DRIVE rather than that same set, outperforming the state-of-the-art with regards to G-mean. Basing the results on G-mean, MCC and F1-scores places the network performance in the middle of the back for DRIVE and STARE, and last for CHASE\_DB1. This trend is not surprising, given deep learning methods performance is dependant on the availability of data to train the system. Compared to the other methods, PixelBNN used 5$\times$ less information for DRIVE, 6.5$\times$ less for STARE, and 18.75$\times$ less information for CHASE\_DB1 (see Table \ref{tab:dataset_stats}). 

\subsection{Computation time} \label{sec:comp_time}
Computation time is a difficult metric to benchmark due to variances in test system components and performance. In an attempt to evaluate this aspect, recent works that share the same GPU - the NVIDIA Titan X - were compared. This is a reasonable comparison as the vast majority of computations are performed on the GPU when training DNNs.  Table \ref{tab:computation_time} shows the comparable methods approximate training and test speeds. Training time was evaluated by normalizing the total time for training the network by the number of training iterations. The total number of iterations was not provided in the multi-classifier article\cite{mo_2017}. Test time is the duration required for evaluating one image at test time. The network evaluated test images in 0.0466s, 8.6$\times$ faster than the state-of-the-art.

\section{Discussion}  \label{sec:discussion} 
Herein, the baseline results for the first known application of PixelBNN, a variant of PixelCNN - a family of FCNs which has never before been applied to fundus images - was evaluated on the task of image segmentation against DRIVE, STARE and CHASE\_DB1 retinal fundus image datasets. Different from the works in the literature, which use cropping and patch segmentation strategies, the proposed method instead resizes the fundus images, shrinking them to 256$\times$256. This incurs a loss of information as many pixels and details are discarded in the process, proportionately reducing the feature space by which the model can learn this task. The decision to use this strategy was primarily driven by computational efficiency, as the methods are intended for use in real time within CAD systems. The cross-testing demonstrates the model's ability to learn generalizable features from each dataset, making it a viable architecture for automated delineation of morphological features within CAD systems. The drop in model performances compared to the state-of-the-art is believed to be caused by the loss of information incurred during preprocessing and will be investigated in future work that also delves into an ablation study. 

\subsection{Conclusion}  \label{sec:conclusion}
This paper proposed a method for segmenting retinal vessels using PixelBNN - a dense multi-stream FCN, using Adam optimization, batch normalization during downsampling and dropout regularization to generate a vessel segmentation mask by converting the feature space of retinal fundus images. F1-score, G-mean and MCC were used to measure network performance, rather than Acc, AUC and $\kappa$. This novel architecture performed well, even after a severe loss of information, even outperforming state-of-the-art methods during cross-testing. This reduction in information also allowed the system to perform 8.5$\times$ faster than the current state-of-the-art at test time, making it a viable candidate for application in real-world CAD systems.  

\section*{Biographies} \label{sec:bios} 
\noindent\textbf{Henry A. Leopold} is a Doctoral candidate at the University of Waterloo in the department of systems design engineering, joint with vision science  through the school of optometry. He received his BASc in materials science bioengineering from the University of Toronto in 2010. 
He has worked in molecular disease research, public health surveillance and entrepreneurial industries. 
His current research interests include artificial intelligence, biomedical informatics, cybernetics and implant engineering. He is a member of SPIE.\\
\noindent\textbf{Jeff Orchard} received degrees in applied mathematics from the University of Waterloo (BMath) and the University of British Columbia (MSc), and received his PhD in Computing Science from Simon Fraser University in 2003. He is an associate professor at the Cheriton School of Computer Science at the University of Waterloo. His research focus is on understanding how the brain works with computational neuroscience,  mathematically modelling and simulating neural networks.\\
\noindent\textbf{John S. Zelek} is an Associate Professor in Systems Design Engineering at the University of Waterloo. He has published over 150 referenced papers and has helped create at least 3 spin off companies resulting from research in his lab. His research interests include autonomous robotics, SLAM, scene \& image understanding, machine learning, augmentative reality, as well as assistive devices. He is also a IEEE member and a registered professional engineer in Ontario Canada.
\\
\noindent\textbf{Vasudevan Lakshminarayanan} is a Professor at University of Waterloo. He received his Ph.D. in Physics from the University of California at Berkeley and is a Fellow of SPIE, OSA, APS, AAAS, etc. He serves on the optics advisory board of the Abdus Salam International Center for Theoretical Physics in Trieste, Italy. He has published widely in various areas including physics, neuroscience, bioengineering, applied math and clinical vision.
\\

\section*{Disclosures}\label{sec:disclosures}
No conflicts of interest, financial or otherwise, are declared by the authors.

\section*{Acknowledgements}\label{sec:acknowledgements}
This work was supported by Discovery and ENGAGE grants from the National Sciences and Engineering Research Council of Canada to Vasudevan Lakshminarayanan. The authors state no conflict of interest and have nothing to disclose.
 
\bibliography{../../../deep_learning_retinal_analysis}
\bibliographystyle{spiejour_modified} 


\listoffigures
\listoftables

\end{document}